\documentclass[runningheads]{llncs}
\usepackage{graphicx}
\usepackage{bm}
\usepackage{tikz}
\usepackage{multirow}
\usepackage{booktabs}
\usepackage{comment}
\usepackage{amsmath,amssymb} 
\usepackage{color}
\usepackage{pifont}
\usepackage{enumitem}
\usepackage{subcaption}
\usepackage[capitalize]{cleveref}

\usepackage[accsupp]{axessibility}  

\usepackage[capitalize]{cleveref}
\crefname{section}{Sec.}{Secs.}
\Crefname{section}{Section}{Sections}
\Crefname{table}{Table}{Tables}
\crefname{table}{Tab.}{Tabs.}

\begin{document}
\pagestyle{headings}
\mainmatter
\def\ECCVSubNumber{1202}  

\title{MetaGait: Learning to Learn an Omni Sample Adaptive Representation for Gait Recognition} 


\titlerunning{MetaGait: Learning to Learn an Omni Sample Adaptive Representation}

%
\author{Huanzhang Dou\inst{1} \and
Pengyi Zhang\inst{1} \and
Wei Su\inst{1} \and \\ Yunlong Yu\inst{2}$^\star$\orcidID{0000-0002-0294-2099}
\and Xi Li\inst{1,3,4}\thanks{Co-corresponding authors.}\orcidID{0000-0003-3023-1662}}

\authorrunning{H. Dou et al.}
%
\institute{College of Computer Science \& Technology, Zhejiang University \\\email{\{hzdou,pyzhang,weisuzju,xilizju\}@zju.edu.cn} \and
College of Information Science \& Electronic Engineering, Zhejiang University\\
\email{\{yuyunlong\}@zju.edu.cn}\\
\and Shanghai Institute for Advanced Study, Zhejiang University \and Shanghai AI Laboratory}
\maketitle

\begin{abstract}

Gait recognition, which aims at identifying individuals by their walking patterns, has recently drawn increasing research attention. However, gait recognition still suffers from the conflicts between the limited binary visual clues of the silhouette and numerous covariates with diverse scales, which brings challenges to the model's adaptiveness.  In this paper, we address this conflict by developing a novel MetaGait that learns to learn an omni sample adaptive representation. Towards this goal, MetaGait injects meta-knowledge, which could guide the model to perceive sample-specific properties, into the calibration network of the attention mechanism to improve the adaptiveness from the omni-scale, omni-dimension, and omni-process perspectives. Specifically, we leverage the meta-knowledge across the entire process, where  Meta Triple Attention and Meta Temporal Pooling are presented respectively to adaptively capture omni-scale dependency from spatial/channel/temporal dimensions simultaneously and to adaptively aggregate temporal information through integrating the merits of three complementary temporal aggregation methods. Extensive experiments demonstrate the state-of-the-art performance of the proposed MetaGait. On CASIA-B, we achieve rank-1 accuracy of 98.7\%, 96.0\%, and 89.3\% under three conditions, respectively. On OU-MVLP, we achieve rank-1 accuracy of 92.4\%.

\end{abstract}

\section{Introduction}
\begin{figure}[t]
\centering
	\subcaptionbox{Diverse scales of covariates.\label{fig:conflict}}{\includegraphics[width = 0.44\textwidth]{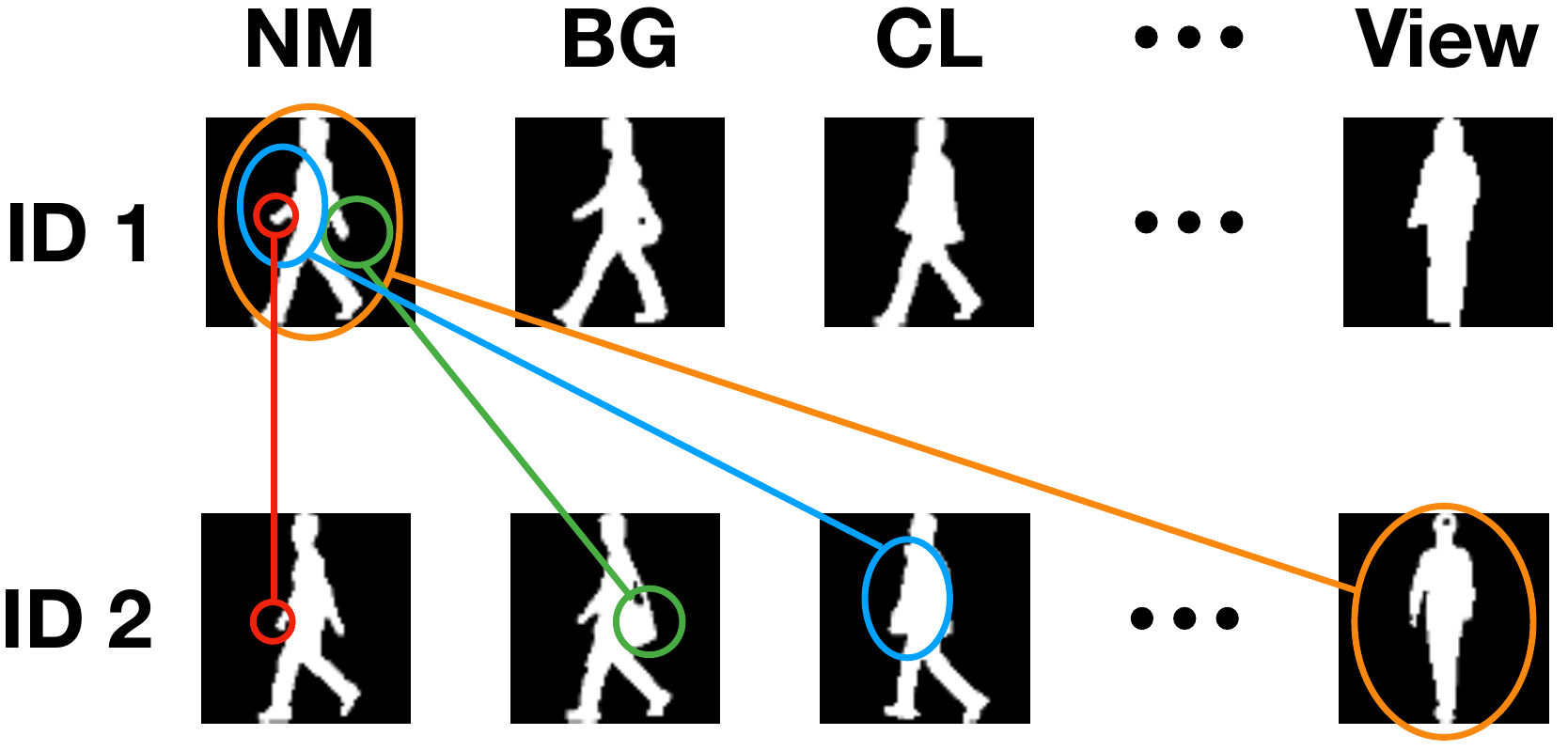}}
	\subcaptionbox{Meta-knowledge parameterization.\label{fig:mhn}}{\includegraphics[width = 0.48\textwidth]{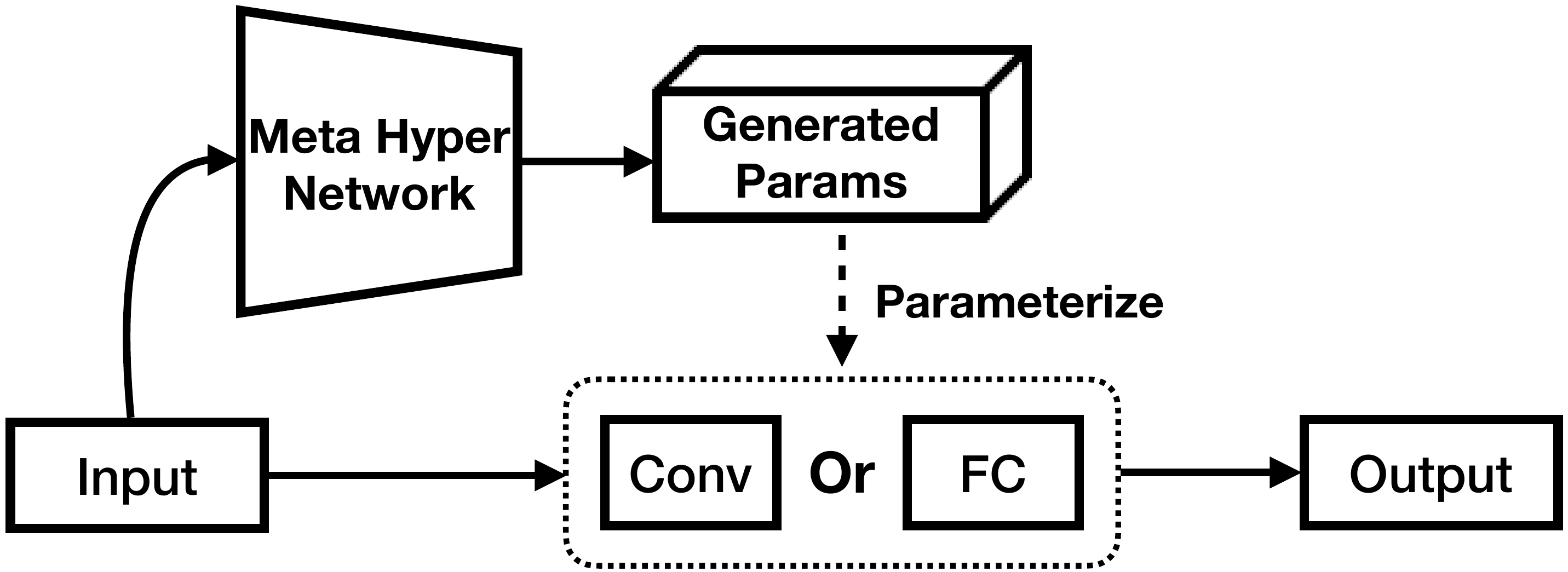}}
	\hfill
\caption{Illustration of the conflicts and the meta-knowledge. \textbf{Left:} The conflicts between limited binary visual clues (colorless and textureless) and numerous covariates with diverse scales, such as bag and clothing, which poses a challenge to the model's adaptiveness. \textbf{Right:} Meta Hyper Network (MHN) learns to learn the meta-knowledge, which could guide the model to perceive sample-specific properties and adaptively parameterize the calibration network.}

\end{figure}

As one of the most promising biometric patterns, gait could be recognized at a long distance without the explicit cooperation of humans, thus having wide applications ranging from security check~\cite{chattopadhyay2014frontal}, video retrieval~\cite{bouchrika2018survey}, to identity identification~\cite{balazia2017human,macoveciuc2019forensic}. Most existing approaches~\cite{Fan_2020_CVPR,lin2020gait,Lin_2021_ICCV} address gait recognition with a two-step process~\cite{sepas2021deep}: feature extraction and temporal aggregation. Though significant advances have been achieved, gait recognition still suffers from the conflict between the  \textit{limited} binary visual clues (colorless and textureless) and \textit{numerous covariates} with diverse scales of the silhouette shown in~\cref{fig:conflict}, which poses a huge challenge to the model's adaptiveness.

Most existing methods tackle this conflict by utilizing the \textit{adaptiveness} of the attention mechanism. For example, the attention mechanism for gait recognition on spatial~\cite{huang2018attention}, channel~\cite{Fan_2020_CVPR}, temporal~\cite{chao2019gaitset},  or two of them \cite{Huang_2021_ICCV} has been effectively explored. However, the existing attention mechanism still has some limitations, which may harm the adaptiveness. First, the calibration network~\cite{hu2018squeeze} that performs feature rescaling in the attention mechanism, is static and limited in capturing dependency at a specific scale. Second, the attention mechanism is applied at most two dimensions while leaving one out. Third, only the feature extraction process is considered, while temporal aggregation is ignored.

To address these limitations, we propose a novel framework called MetaGait, to enhance the adaptiveness of the attention mechanism for gait recognition from three perspectives: \textit{omni scale}, \textit{omni dimension}, and \textit{omni process}. The core idea of MetaGait is to leverage \textit{meta-knowledge}~\cite{hospedales2020meta,zhang2017supplementary,cheng2019swiftnet}, which could guide the model to perceive sample-specific properties, into the calibration network of attention mechanism. Specifically, the meta-knowledge is learned by a Meta Hyper Network (MHN) shown in~\cref{fig:mhn} and MHN could parameterize the calibration network in a sample adaptive manner instead of being fixed.

Specifically, benefited from the meta-knowledge, we first present Meta Triple Attention (MTA) to adaptively capture the omni-scale dependency in the feature extraction process, leading to the ability to extract walking patterns from diverse scales. The calibration network of MTA is achieved by a weighted dynamic multi-branch structure with diverse receptive fields and parameterized by the meta-knowledge. Second,  MTA is designed in homogeneous and applied on spatial/channel/temporal dimensions simultaneously. Third, apart from the feature extraction process, we present Meta Temporal Pooling (MTP) on temporal aggregation for adaptively integrating temporal information. MTP leverages the meta-knowledge to parameterize an attention-based weighting network, which could excavate the relation between three mainstream temporal aggregation methods with complementary properties (\textit{i.e.,} Max/Average/GeM Pooling~\cite{Lin_2021_ICCV}). Therefore, MTP could adaptively aggregate their merits for comprehensive and discriminative representation.

Extensive experiments are conducted on two widely used datasets to evaluate the proposed MetaGait framework. The superior results demonstrate that MetaGait outperforms other state-of-the-art methods by a considerable margin, which verifies its effectiveness and adaptiveness.

The major contributions of this work are summarized as follows:

\begin{itemize}
    \item We present MetaGait framework to address the conflict between limited binary visual clues and numerous covariates with diverse scales. The core idea is to introduce the meta-knowledge learned from Meta Hyper Network to enhance the calibration network's adaptiveness in the attention mechanism.
	\item We present Meta Triple Attention (MTA) for the feature extraction process, which aims at adaptively capturing the omni-scale dependency on spatial, channel, and temporal dimensions simultaneously.
	\item We present attention-based Meta Temporal Pooling (MTP), which could adaptively integrate the merits of three temporal aggregation methods with complementary properties in the temporal aggregation process.

\end{itemize}

\label{sec:intro}
\section{Related Work}

\subsection{Gait Recognition}

\textbf{Model-based Approaches.} These methods~\cite{6117582,bodor2009view,4378964,1613073,liao2020model,kastaniotis2016pose} aim at modeling the structure of human body from pose information~\cite{cao2017realtime,Sun_2019_CVPR}. For example, Wang \textit{et al.}~\cite{wang2004fusion} propose to use the angle change of body joints to model the walking pattern of different individuals. The advantage of these methods is that they are robust to the clothing and viewpoints conditions. Nevertheless, the model-based approaches suffer from expensive computational costs, accurate pose estimation results, missing ID-related shape information, and extra data collection devices.

\noindent\textbf{Appearance-based Approaches.} These methods~\cite{5299188,6680737,10.1007/11744078_12,1699873,dou2021versatilegait,1561189,5522296,samangooei2010performing,6478807,dou2022gaitmpl,Dou_2023_CVPR}  learn the features from the silhouette sequences without explicitly modeling the human body structure. For example,  GaitSet~\cite{chao2019gaitset} and GLN~\cite{hou2020gait} deem each silhouette sequence as an unordered set for recognition. GaitPart~\cite{Fan_2020_CVPR} utilizes 1D convolutions to extract temporal information and aggregate it by a summation or a concatenation. MT3D~\cite{lin2020gait} and 3DLocal~\cite{Huang_2021_ICCV2} propose to exploit 3D convolutions to extract spatial and temporal information at the same time. Appearance-based approaches become popular for their flexibility, conciseness, and effectiveness. The proposed MetaGait is in the scope of appearance-based gait recognition.

\noindent\textbf{Attention Mechanism.} Visual attention~\cite{hu2018squeeze,he2020gta,Wang_2020_CVPR,woo2018cbam,zhang2022adaptive}, which highlights informative clues and suppresses useless ones, has drawn research attention, and it has been applied to gait recognition successfully. GaitPart~\cite{Fan_2020_CVPR} performs short-range modeling by channel attention. Zhang \textit{et al.}~\cite{Zhang2019} introduce temporal attention to learn the attention score of each frame by LTSM~\cite{du2015hierarchical}. Besides, there are methods~\cite{huang2018attention,8643814} that apply spatial attention. In this paper, we propose to alleviate the conflict between limited visual clues and various covariates with diverse scales from the perspective of the attention mechanism's adaptiveness.

\subsection{Dynamic Networks} Dynamic networks can adjust the structures/parameters in an input-dependent manner, leading to several advantages like efficiency, representation power, adaptiveness, and generalizability. Dynamic networks can be mainly divided into dynamic architectures~\cite{wang2018skipnet,veit2018convolutional,ma2018modeling,eigen2013learning,bengio2015conditional,jacobs1991adaptive,lin2017runtime} and dynamic parameters~\cite{yang2019condconv,chen2020dynamic,harley2017segmentation,su2019pixel,dai2017deformable,gao2019deformable,shan2020meta,denil2013predicting,Su_2023_CVPR}. SkipNet~\cite{wang2018skipnet} and conv-AIG~\cite{veit2018convolutional} are two representative approaches to enabling layer skipping to control the architecture. CondConv~\cite{yang2019condconv} utilizes the weighted sum of the candidate convolutions according to the input.

Further, Zhang~\textit{et al.}~\cite{zhang2017supplementary} point out that dynamic network can be seen as a form of meta-learning~\cite{antoniou2018train,hospedales2020meta,devos2019reproducing,finn2019online,zhang2017supplementary} in learning to learn fashion. In this paper, we leverage the dynamic network for the first time to inject meta-knowledge into the calibration network for improving the model's adaptiveness.

 \begin{figure}[t]
 \small
    \begin{center}
       \includegraphics[width=0.95\textwidth]{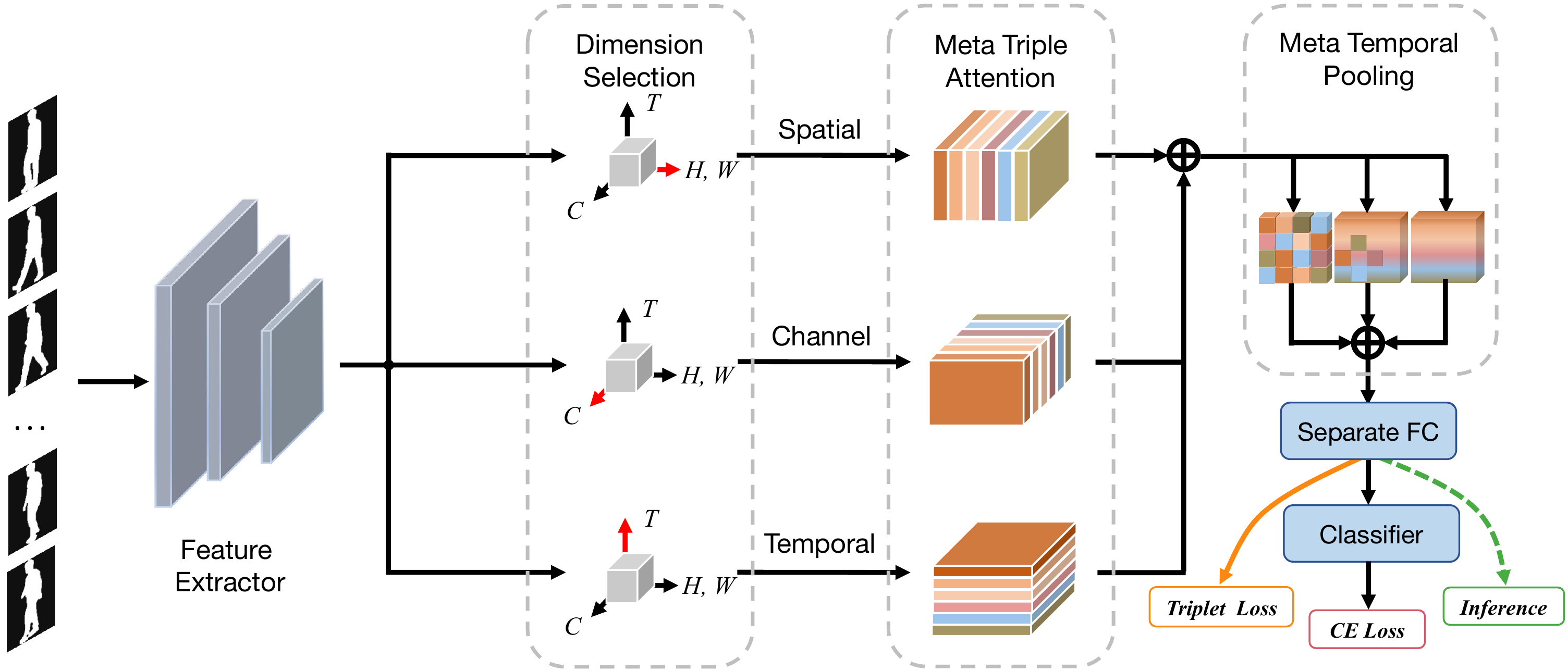}
    \end{center}
    \caption{Overview of MetaGait. Dimension selection refers to the transpose operation for Meta Triple Attention on the selected dimension. Meta Temporal Pooling adaptively aggregates the merits of three temporal aggregation methods with complementary properties. Separate FC is followed by~\cite{chao2019gaitset,Lin_2021_ICCV}.}
    \label{fig:overview}
 \end{figure}

\section{Method}
In this section, we first present the overview of MetaGait in~\cref{fig:overview} and then elaborate on the meta-knowledge learned by Meta Hyper Network (MHN). Further, we introduce two modules that use meta-knowledge in two separate processes, \textit{i.e.,} Meta Triple Attention for feature extraction and Meta Temporal Pooling for temporal aggregation. Finally, the details of the optimization are described.

\subsection{Overview}

The overview of MetaGait is shown in~\cref{fig:overview}. First, the gait sequences are fed into the feature extractor, and the feature maps are transposed to perform Meta Triple Attention, which models the omni-scale representation on spatial/channel/temporal dimensions simultaneously. Then, Meta Temporal Pooling adaptively integrates the temporal information with three complementary temporal aggregation methods. Finally, the final objective is computed by the features from separated fully-connected layer~\cite{chao2019gaitset,Lin_2021_ICCV}.

\subsection{Meta Hyper Network}
Considering the fact that most attention mechanism in gait recognition applies a static strategy to their calibration network~\cite{hu2018squeeze}, which may harm the model's adaptiveness, we propose Meta Hyper Network (MHN) to parameterize the calibration network of the attention mechanism adaptively.  As shown in~\cref{fig:mhn}, MHN learns information on data-specific properties of input gait silhouette sequences, \textit{i.e.,} \textit{meta-knowledge}, and generates the parameters of calibration network in a sample adaptive manner.

Given the input $X\in \mathbb{R}^{C\times T\times H\times W}$, let $\bm{F}(\cdot)$ be a mapping network,  the key to MHN is learning a mapping $\bm{F}_{meta}$ from $\mathbb{R}^C$ to $\mathbb{R}^{N}$ that is used to parameterize the calibration network $\bm{F}_{cali}$ with its parameters $\bm{W}_{cali}$, \textit{i.e.,} fully connected layer ($N=C^{\prime}\times C $) or convolution ($N=C^{\prime}\times C\times k_h\times k_w \times k_t$). $C$, $C^{\prime}$, and $k$ are the input channel, output channel, and kernel size, respectively. Therefore, the attention mechanism with the meta-knowledge can be formulated as:
\begin{equation}
    \bm{f} = \bm{F}_{cali}(X)\otimes X, \,\,\,\,\,\,\,    s.t. \,\, \bm{W}_{cali}=\bm{F}_{meta}(X),
\end{equation}
where $\otimes$ is element-wise multiplication. Specifically, MHN first utilizes Global Averange Pooling (GAP) on spatial and temporal dimensions to computes the statics $m\in \mathbb{R}^{C\times 1\times 1\times 1}$ for MHN as: 
\begin{equation}
    m= GAP(X) = \frac{1}{H\times W\times T}\sum^{H}_{i=1}\sum^W_{j=1}\sum^T_{k=1}X(i,j,k).
    \label{eq:gap}
\end{equation}

Then, the meta-knowledge $\bm{W}_{meta}=\{\bm{W}_{meta_1}\in \mathbb{R}^{C\times C}, \bm{W}_{meta_2}\in\mathbb{R}^{N\times C}\}$ learned by MHN generates the sample adaptive parameters $\bm{W}_{cali}$ of the calibration network by a Multi-Layer Perceptron (MLP) with Leaky ReLU $\delta$ as:
\begin{equation}
   \bm{W}_{cali}=\delta(\bm{W}_{meta_2}\delta(\bm{W}_{meta_1}m).
\end{equation}

In this paper, the meta-knowledge is used to improve the adaptiveness of the attention mechanism on the modules as follows: the global/local calibration stream in Meta Triple Attention (MTA), the soft aggregation gate in MTA, and the weighting network of Meta Temporal Pooling described in~\cref{sec:MTP}.

\subsection{Meta Triple Attention}
Though previous attention methods in feature extraction achieve great success, they mainly suffer from two issues. First, they could only capture fixed-scale dependency while numerous covariates present diverse scales, which may harm the model's adaptiveness. For example, the covariates with small visual changes like bag carrying only require a small receptive field while a large one would bring noises. In contrast, the covariates with significant visual changes like viewpoints require a large receptive field while the small one cannot cover complete visual changes. Second, they only perform attention on two dimensions at most while leaving one dimension out, which is ineffective and insufficient.

To enhance the adaptiveness of the attention mechanism in the feature extraction process, we propose  Meta Triple Attention (MTA), which injects the meta-knowledge into its feature rescaling and feature aggregation to capture omni-scale dependency and perform the omni-dimension attention mechanism sufficiently. Thus, MTA differs from the previous attention mechanism in two corresponding aspects: 1) MTA could cope with the numerous covariates at omni scale; 2) MTA performs homogeneous attention mechanism on spatial, channel, and temporal dimensions simultaneously rather than one or two of them. Note that we describe MTA in channel attention on each frame for simplicity while applied in all three dimensions in practice.

\begin{figure}[t]
\centering
	\subcaptionbox{Meta Triple Attention.\label{fig:mta}}{\includegraphics[width = 0.63\textwidth]{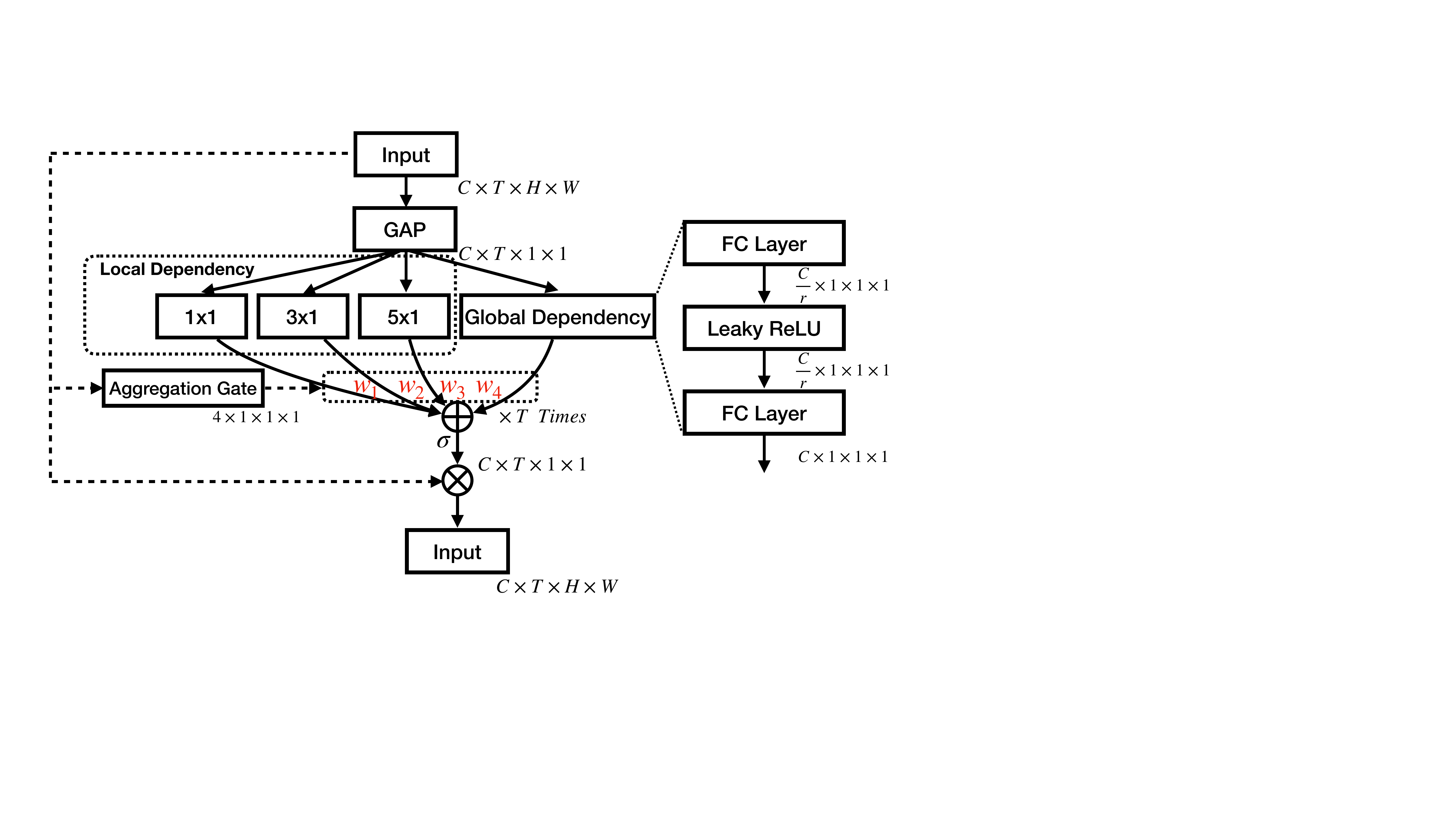}}
	\subcaptionbox{Omni-scale representation.\label{fig:omniscale}}{\includegraphics[width = 0.35\textwidth]{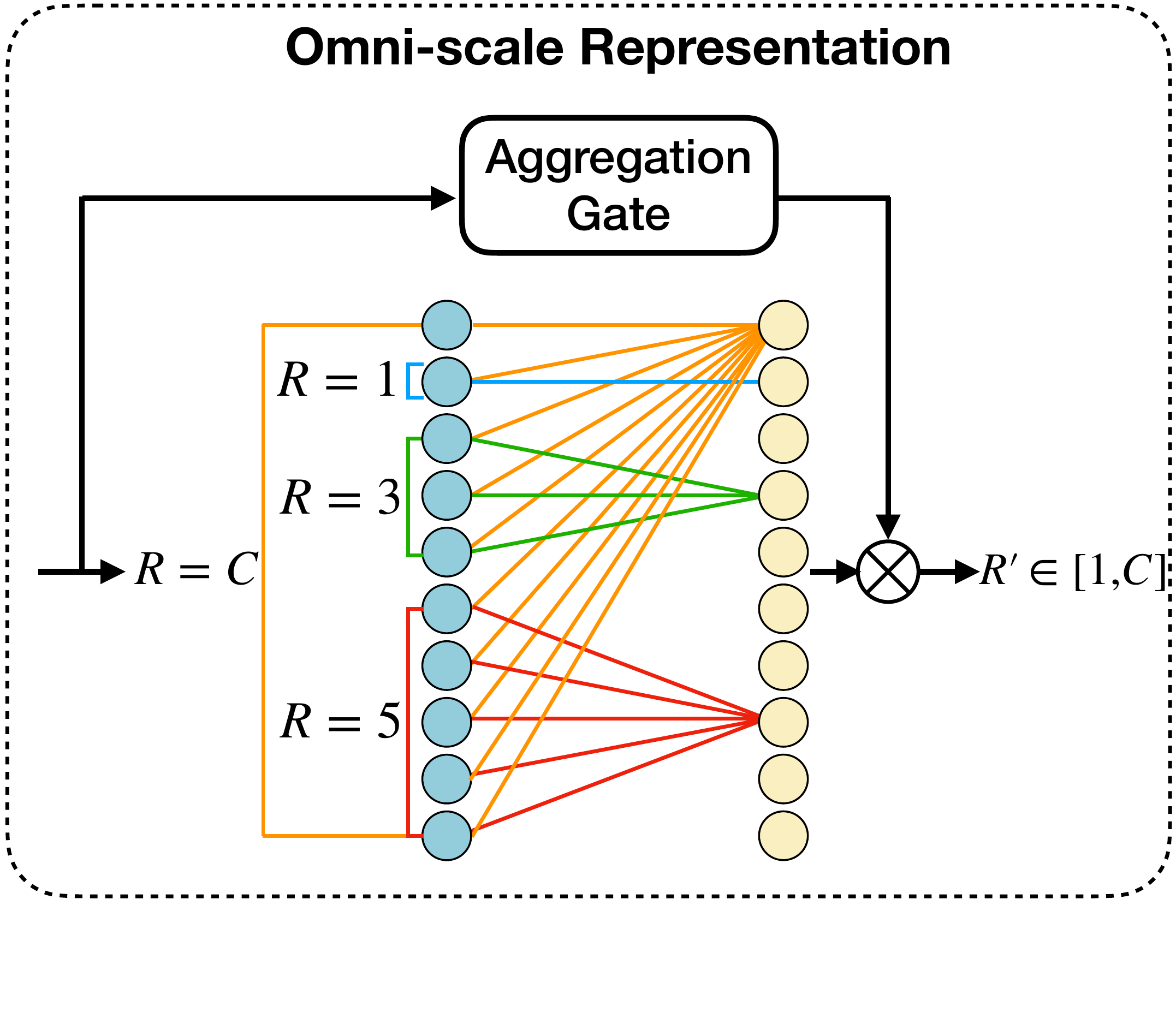}}
	\hfill
\caption{Illustration of Meta Triple Attention (MTA) and omni-scale representation. (a) MTA is composed of a multi-branch structure with diverse receptive fields weighted by the aggregation gate. (b) MTA achieves omni-scale representation via adaptively weighting the multi-branch structure by soft aggregation gate, leading to the outputs's receptive field $\bm{R}^{\prime}\in[1,C]$.}

\end{figure}

Specifically, we leverage the global and local dependency modeling in frame-level with the weighted multi-branch structure to achieve omni-scale representation. As shown in~\cref{fig:mta}, for the global channel dependency relation modeling in frame level, a GAP is first applied on the spatial to obtain each frame's statistics $s\in \mathbb{R}^{C\times 1\times 1\times 1}$. Then, to effectively capture dimension-wise non-linear global dependency $\bm{f}_{global}$ and evaluate the channel-wise importance, MTA utilizes an MLP activated by Leaky ReLu, which follows the bottleneck design~\cite{hu2018squeeze,he2016deep} with a dimension reduction ratio $r$:
\begin{equation}
    \bm{f}_{global} = \bm{F}_{global}(s) = \bm{W}_{g_2}\delta(\bm{W}_{g_1}s),
    \label{eq:global}
\end{equation}
where the parameters $\bm{W}_g=\{\bm{W}_{g_1}\in \mathbb{R}^{\frac{C}{r}\times C}, \bm{W}_{g_2}\in \mathbb{R}^{C\times\frac{C}{r}}\}$ of global calibration stream  $\bm{F}^{global}_{cali}$ is adaptively parameterized by MHN.

For local dependency modeling of the calibration network, we design a multi-branch convolutional structure with diverse receptive fields (\textit{i.e.,} kernel size). Therefore, each local calibration stream could capture dependency at a specific scale. To learn omni-scale representation, we propose to aggregate the output $\bm{f}_{global}$ and $\bm{f}_{local}$ of global and local streams in a sample adaptive manner as~\cref{eq:attention1}  instead of being fixed, which is achieved by a soft aggregation gate $\bm{G}$ with meta-knowledge. Next, Sigmoid $\sigma$ is applied to mapping the values of the attention vector into $[0,1]$:

\begin{equation}
    \bm{f}_{mta} = \sigma(\bm{G}(s)[L+1] * \bm{f}_{global} + \sum_{l=1}^{L}\bm{G}(s)[l] * \bm{f}^l_{local})\otimes X, \,\,\,    s.t. \,\, L\geq 1,
    \label{eq:attention1}
\end{equation}
where $L$ denotes the number of the receptive field sizes in local stream of the calibration network. The output of the soft aggregation gate $\bm{G}$ is a vector with $\mathbb{R}^{L+1}$ to weight each stream according to the input. Specifically, $\bm{G}$ is implemented by GAP, an MLP mapping from $\mathbb{R}^{C\times 1\times1 \times 1}$ to $\mathbb{R}^{(L+1)\times 1\times1 \times 1}$, and Sigmoid in sequential. Therefore, the output's receptive field $\bm{R}^{\prime}$ is adaptively ranging from $1$ to global receptive field $C$, which could capture omni-scale dependency.

Besides, previous approaches design heterogeneous attention modules for each dimension to fit the dependency scale that each dimension needs to model. Benefited from the omni-scale dependency modeling, MTA can be efficiently performed on three different dimensions in homogeneous.
\label{mhn}
\subsection{Meta Temporal Pooling}

To achieve omni-process sample adaptive representation, the meta-knowledge is injected into the temporal aggregation apart from feature extraction. In the recent gait literature~\cite{chao2019gaitset,Fan_2020_CVPR,Lin_2021_ICCV}, Global Max Pooling (GMP), Global Average Pooling (GAP), and GeM Pooling~\cite{radenovic2018fine} along the temporal dimension are the mainstream temporal aggregation methods, which represent the salient information, overall information, and an intermediate form between the former two methods, respectively. They can be formulated as:

\begin{subequations}

\begin{align}
 \bm{Max}(\cdot) = Pool_{Max}^{T\times 1\times 1}(\cdot), \label{Za}\\
\bm{Mean}(\cdot) = Pool_{Avg}^{T\times 1\times 1}(\cdot), \label{Zb} \\
\bm{GeM}(\cdot) = (\bm{Mean}(\cdot)^p)^{\frac{1}{p}}, \label{Zc} 
\end{align}
\end{subequations}
where $p$ in GeM Pooling is a learnable parameter. Though these temporal aggregation methods have been validated their effectiveness individually, their relation is under-explored. We argue that different temporal aggregation methods have their own merits and complementarities to each other, which can be excavated to adaptively integrated temporal clues according to the properties of inputs. Specifically, GMP preserves the most salient information along the temporal dimension while ignoring the majority of information. By contrast, GAP includes the overall temporal information, but the salient information would be diluted out. Though GeM, an intermediate form, can obtain salient temporal information while preserving overall one, it is less robust than GAP and GMP due to the learning stability of unconstrained learnable parameter $p$.
 \begin{figure}[t]
 \small
    \begin{center}
       \includegraphics[width=0.65\textwidth]{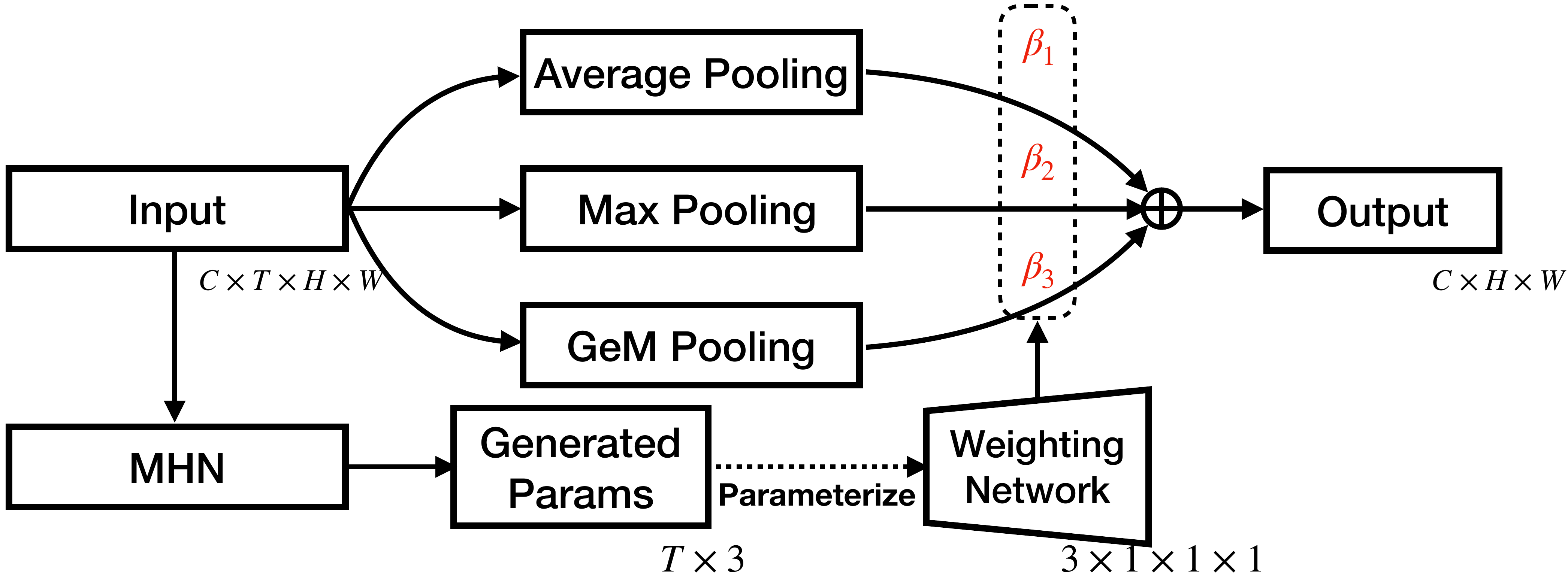}
    \end{center}
    \caption{The illustration of Meta Temporal Pooling (MTP), which aims at leveraging meta-knowledge from MHN to adaptively integrate the merits from three complementary temporal aggregation methods.}
    \label{fig:aggregation}
 \end{figure}

To fully exploit their merits, we leverage the meta-knowledge learned by MHN to adaptively integrate the features produced by three temporal aggregation methods as shown in~\cref{fig:aggregation}. In detail, we first compute the statics using GAP on spatial and channel dimensions, and we utilize MHN to generate the parameter ($N=3T$), which is used to parameterize the weighting network, \textit{i.e.,} an FC layer with $\bm{W}_t\in \mathbb{R}^{3\times T}$ followed by a Sigmoid. Therefore, the weights of different temporal aggregation methods $\beta \in \mathbb{R}^3$ can be obtained as:
\begin{equation}
    \beta = \sigma(\bm{W}_{t}(GAP(\bm{f}_{mta}))).
    \label{eq:beta}
\end{equation}
Then, $\beta$ adaptively weights the features of three complementary temporal aggregation methods and obtain omni sample adaptive representation $\bm{f}_{omni}$ as:

\begin{equation}
    \bm{f}_{omni} = \beta_1\bm{Mean}(\bm{f}_{mta}) + \beta_2\bm{Max}(\bm{f}_{mta}) + \beta_3\bm{GeM}(\bm{f}_{mta})
    \label{eq:beta1}
\end{equation}

\label{sec:MTP}
\subsection{Optimization}
Following the optimization strategy\cite{Lin_2021_ICCV,hou2020gait,Huang_2021_ICCV}, we apply Triplet Loss~\cite{hermans2017defense} $\bm{\mathcal{L}}_{tri}$ and Cross-Entropy loss $\bm{\mathcal{L}}_{ce}$ on each horizontal feature independently to train our model as~\cref{eq:total}. The similarity metric is set to Euclidean distance.

\begin{equation}
    \bm{\mathcal{L}}_{total} = \bm{\mathcal{L}}_{tri} + \bm{\mathcal{L}}_{ce}.
    \label{eq:total}
\end{equation}

\section{Experiments}
\subsection{Datasets}
\noindent\textbf{CASIA-B}~\cite{1699873}\textbf{.} It is composed of 124 IDs, each of which has 10 groups of sequences, \textit{i.e.,} 6 normal walking (NM), 2 walking with a bag (BG), 2 walking in coats (CL). The views are uniformly distributed in $[0^{\circ}, 180^{\circ}]$. For evaluation, the protocol is adopted as~\cite{chao2019gaitset}, \textit{i.e.,} small-scale training (ST), medium-scale training (MT), and large-scale training (LT). These three settings select the first 24/62/74 IDs as the training set and the rest 100/62/50 IDs as the test set, respectively. During the evaluation, the first four sequences of each ID under NM are deemed as the gallery, and the rest are used as the probe.

\noindent\textbf{OU-MVLP}~\cite{takemura2018multi}\textbf{.} It is the largest dataset consisting of 10,307 IDs. In OU-MVLP, there are 1 waling condition (NM) with 2 sequences and 14 views, which are uniformly distributed between $[0^{\circ}, 90^{\circ}]$ and $[180^{\circ}, 270^{\circ}]$. The training set and test set are composed of 5,153 IDs and 5,154 IDs, respectively. For evaluation, the first sequence of each ID is adopted as the gallery, and the rest is the probe.

\subsection{Implementation Details}
\noindent\textbf{Hyper-parameters.} 1) The resolution of the silhouette is resized to $64\times44$ or $128\times88$ following~\cite{hou2020gait,Huang_2021_ICCV2,Huang_2021_ICCV}; 2) In a mini-batch, the number of the IDs and the sequences of each ID is set to (8, 8) for CASIA-B and (32, 8) for OU-MVLP; 3) Adam optimizer is used with a learning rate of 1\textit{e}-4; 4) We train our model for 100k iterations for CASIA-B and 250k for OU-MVLP, where the learning rate is reduced to 1\textit{e}-5 at 150k iterations; 5) The margin of Triplet loss is set to 0.2; 6) The reduction ratio $r$ in this paper is all set to 2.

\noindent\textbf{Training Details.} 1) The feature extractor is following the global and local backbone in~\cite{Lin_2021_ICCV}; 2) The channels of the feature extractor in the three stages are set to (32, 64, 128) for CASIA-B and double for OU-MVLP. 3) The local stream of MTA is implemented with Conv1d and Conv2d for channel/temporal and spatial dimensions, respectively. The receptive fields of the local stream are set to \{1,3,5\}. Refer to supplementary materials for more details.

\subsection{Comparison with State-of-the-Art Methods}
\begin{table*}[!ht]
    \centering
    \caption{Averaged rank-1 accuracy on CASIA-B, excluding identical views cases.}
    \setlength{\tabcolsep}{0.2mm}
    \renewcommand{\arraystretch}{0.7}
   \begin{tabular}{l|c|c|ccccccccccc|c}
    \toprule  
    \multicolumn{3}{l|}{Gallery NM \#1-4}                             & \multicolumn{11}{c|}{0$^{\circ}$-180$^{\circ}$}                                                & \multirow{2}{*}{Mean} \\ \cmidrule{1-14}
    Prob.                & \multicolumn{1}{c|}{Res.} & Method   & 0$^{\circ}$    & 18$^{\circ}$   & 36$^{\circ}$   & 54$^{\circ}$   & 72$^{\circ}$   & 90$^{\circ}$   & 108$^{\circ}$  & 126$^{\circ}$  & 144$^{\circ}$  & 162$^{\circ}$  & 180$^{\circ}$  &                       \\ \hline
    \multirow{12}{*}{NM} & \multirow{7}{*}{$64 \times 44$}          & GaitSet  & 90.8 & 97.9 & 99.4 & 96.9 & 93.6 & 91.7 & 95.0 & 97.8 & 98.9 & 96.8 & 85.8 & 95.0                  \\ \cline{15-15} 
                         &                                 & GaitPart & 94.1 & 98.6 & 99.3 & 98.5 & 94.0 & 92.3 & 95.9 & 98.4 & 99.2 & 97.8 & 90.4 & 96.2                  \\ \cline{15-15} 
                         &                                 & MT3D     & 95.7 & 98.2 & 99.0 & 97.5 & 95.1 & 93.9 & 96.1 & 98.6 & 99.2 & 98.2 & 92.0 & 96.7                  \\ \cline{15-15} 
                         &                                 & CSTL     & 97.2 & 99.0 & 99.2 & 98.1 & 96.2 & 95.5 & \textbf{97.7} & 98.7 & 99.2 & 98.9 & 96.5 & 97.8                  \\ \cline{15-15} 
                         &                                 & 3DLocal  & 96.0 & 99.0 & 99.5 &98.9 & 97.1 & 94.2 & 96.3 & 99.0 & 98.8 & 98.5 & 95.2 & 97.5                  \\ \cline{15-15} 
                         &                                 & GaitGL &      96.0 &98.3& 99.0& 97.9& 96.9 &95.4& 97.0& 98.9& 99.3& 98.8& 94.0& 97.4                      \\ \cline{15-15} 
                         &                                 & \textbf{MetaGait}     & \textbf{97.3}&\textbf{99.2}&\textbf{99.5}&\textbf{99.1}&\textbf{97.2}&\textbf{95.5}&97.6&\textbf{99.1}&\textbf{99.3}&\textbf{99.1}&\textbf{96.7}&\textbf{98.1}  \\ \cline{2-15} 
                         & \multirow{5}{*}{$128 \times 88$}        & GaitSet  & 91.4 & 98.5 & 98.8 & 97.2 & 94.8 & 92.9 & 95.4 & 97.9 & 98.8 & 96.5 & 89.1 & 95.6                  \\ \cline{15-15} 
                         &                                 & GLN      & 93.2 & 99.3 & 99.5 & 98.7 & 96.1 & 95.6 & 97.2 & 98.1 & 99.3 & 98.6 & 90.1 & 96.9                  \\ \cline{15-15} 
                         &                                 & CSTL     & 97.8 & 99.4 & 99.2 & 98.4 & 97.3 & 95.2 & 96.7 & 98.9 & 99.4 & 99.3 & 96.7 & 98.0                  \\ \cline{15-15} 
                         &                                 & 3DLocal  & 97.8 & 99.4 & 99.7 & 99.3 & 97.5 & 96.0 & 98.3 & 99.1 & 99.9 & 99.2 & 94.6 & 98.3                  \\ \cline{15-15} 
                         &                                 & \textbf{MetaGait}     &     \textbf{98.1}&\textbf{99.4}&\textbf{99.8}&\textbf{99.4}&\textbf{97.6}&\textbf{96.7}&\textbf{98.5}&\textbf{99.3}&\textbf{99.9}&\textbf{99.6}&\textbf{97.0}&\textbf{98.7}                \\ \hline
    \multirow{12}{*}{BG} & \multirow{7}{*}{$64 \times 44$}          & GaitSet  & 83.8 & 91.2 & 91.8 & 88.8 & 83.3 & 81.0 & 84.1 & 90.0 & 92.2 & 94.4 & 79.0 & 87.2                  \\ \cline{15-15} 
                         &                                 & GaitPart & 89.1 & 94.8 & 96.7 & 95.1 & 88.3 & 84.9 & 89.0 & 93.5 & 96.1 & 93.8 & 85.8 & 91.5                  \\ \cline{15-15} 
                         &                                 & MT3D     & 91.0 & 95.4 & 97.5 & 94.2 & 92.3 & 86.9 & 91.2 & 95.6 & 97.3 & 96.4 & 86.6 & 93.0                  \\ \cline{15-15} 
                         &                                 & CSTL     & 91.7 & 96.5 & 97.0 & 95.4 & 90.9 & 88.0 & 91.5 & 95.8 & 97.0 & 95.5 & 90.3 & 93.6                  \\ \cline{15-15} 
                         &                                 & 3DLocal  & 92.9 & 95.9 & \textbf{97.8} & 96.2 & 93.0 & 87.8 & 92.7 & 96.3 & 97.9 & 98.0 & 88.5 & 94.3                  \\ \cline{15-15} 
                         &                                 & GaitGL   &    92.6 &96.6& 96.8 &95.5& 93.5& 89.3& 92.2& 96.5& 98.2& 96.9& 91.5& 94.5                   \\ \cline{15-15} 
                         &                                 & \textbf{MetaGait}     &    \textbf{92.9}&\textbf{96.7}&97.1&\textbf{96.4}&\textbf{94.7}&\textbf{90.4}&\textbf{92.9}&\textbf{97.2}&\textbf{98.5}&\textbf{98.1}&\textbf{92.3}&\textbf{95.2}                 \\ \cline{2-15} 
                         & \multirow{5}{*}{$128 \times 88$}        & GaitSet  & 89.0 & 95.3 & 95.6 & 94.0 & 89.7 & 86.7 & 89.7 & 94.3 & 95.4 & 92.7 & 84.4 & 91.5                  \\ \cline{15-15} 
                         &                                 & GLN      & 91.1 & 97.7 & 97.8 & 95.2 & 92.5 & 91.2 & 92.4 & 96.0 & 97.5 & 95.0 & 88.1 & 94.0                  \\ \cline{15-15} 
                         &                                 & CSTL     & 95.0 & 96.8 & 97.9 & 96.0 & 94.0 & 90.5 & 92.5 & 96.8 & 97.9 & \textbf{99.0} & \textbf{94.3} & 95.4                  \\ \cline{15-15} 
                         &                                 & 3DLocal  & 94.7 & 98.7 & 98.8 & 97.5 & 93.3 & 91.7 & 92.8 & 96.5 & 98.1 & 97.3 & 90.7 & 95.5                  \\ \cline{15-15} 
                         &                                 & \textbf{MetaGait}     &   \textbf{95.1}&\textbf{98.9}&\textbf{99.0}&\textbf{97.8}&\textbf{94.0}&\textbf{92.0}&\textbf{92.9}&\textbf{96.9}&\textbf{98.2}&98.4&93.5&\textbf{96.0}                 \\ \hline
    \multirow{12}{*}{CL} & \multirow{7}{*}{$64 \times 44$}          & GaitSet  & 61.4 & 75.4 & 80.7 & 77.3 & 72.1 & 70.1 & 71.5 & 73.5 & 73.5 & 68.4 & 50.0 & 70.4                  \\ \cline{15-15} 
                         &                                 & GaitPart & 70.7 & 85.5 & 86.9 & 83.3 & 77.1 & 72.5 & 76.9 & 82.2 & 83.8 & 80.2 & 66.5 & 78.7                  \\ \cline{15-15} 
                         &                                 & MT3D     & 76.0 & 87.6 & 89.8 & 85.0 & 81.2 & 75.7 & 81.0 & 84.5 & 85.4 & 82.2 & 68.1 & 81.5                  \\ \cline{15-15} 
                         &                                 & CSTL     & 78.1 & 89.4 & 91.6 & 86.6 & 82.1 & 79.9 & 81.8 & 86.3 & 88.7 & 86.6 & 75.3 & 84.2                  \\ \cline{15-15} 
                         &                                 & 3DLocal  & 78.2 & 90.2 & 92.0 & 87.1 & 83.0 & 76.8 & 83.1 & 86.6 & 86.8 & 84.1 & 70.9 & 83.7                  \\ \cline{15-15} 
                         &                                 & GaitGL   &     76.6& 90.0& 90.3& 87.1& 84.5& 79.0& 84.1& 87.0& 87.3& 84.4& 69.5& 83.6                   \\ \cline{15-15} 
                         &                                 & \textbf{MetaGait}     & \textbf{80.0}&\textbf{91.8}&\textbf{93.0}&\textbf{87.8}&\textbf{86.5}&\textbf{82.9}&\textbf{85.2}&\textbf{90.0}&\textbf{90.8}&\textbf{89.3}&\textbf{78.4}&\textbf{86.9}          \\ \cline{2-15} 
                         & \multirow{5}{*}{$128 \times 88$}        & GaitSet  & 66.3 & 79.4 & 84.5 & 80.7 & 74.6 & 73.2 & 74.1 & 80.3 & 79.7 & 72.3 & 62.9 & 75.3                  \\ \cline{15-15} 
                         &                                 & GLN      & 70.6 & 82.4 & 85.2 & 82.7 & 79.2 & 76.4 & 76.2 & 78.9 & 77.9 & 78.7 & 64.3 & 77.5                  \\ \cline{15-15} 
                         &                                 & CSTL     & 84.1 & 92.1 & 91.8 & 87.2 & 84.4 & 81.5 & 84.5 & 88.4 & 91.6 & 91.2 & 79.9 & 87.0                  \\ \cline{15-15} 
                         &                                 & 3DLocal  & 78.5 & 88.9 & 91.0 & 89.2 & 83.7 & 80.5 & 83.2 & 84.3 & 87.9 & 87.1 & 74.7 & 84.5                  \\ \cline{15-15} 
                         &                                 & \textbf{MetaGait}     &    \textbf{87.8}&\textbf{94.6}&\textbf{93.5}&\textbf{90.3}&\textbf{87.1}&\textbf{84.3}&\textbf{86.1}&\textbf{89.7}&\textbf{93.9}&\textbf{93.4}&\textbf{81.7}&\textbf{89.3}     \\ \bottomrule          
    \end{tabular}
    \label{tab:casia}
    \end{table*}

\noindent\textbf{Results on CASIA-B.} To evaluate MetaGait on cross-view and large resolution scenarios, we conduct a comparison between MetaGait and latest SOTA as shown in~\cref{tab:casia}, where MetaGait outperforms SOTA at most views and both two resolutions. Specifically, under NM/BG/CL conditions, MetaGait outperforms previous methods by \textbf{0.3\%}/\textbf{0.4\%}, \textbf{0.7\%}/\textbf{0.5\%}, and \textbf{2.7\%}/\textbf{2.3\%} at the resolution of $64\times 44$/$128\times 88$ \textbf{at least}. Further, MetaGait achieves rank-1 accuracies over \textbf{98\%} and \textbf{96\%} under NM and BG, respectively. More importantly, the considerable performance gain on the most challenging condition CL narrows the gap between the performance of NM and CL to less than \textbf{10\%}, which verifies the robustness of MetaGait under the cross-walking-condition scenario.

\begin{figure}[!t]
\centering
	\subcaptionbox{NM}{\includegraphics[width = 0.24\textwidth]{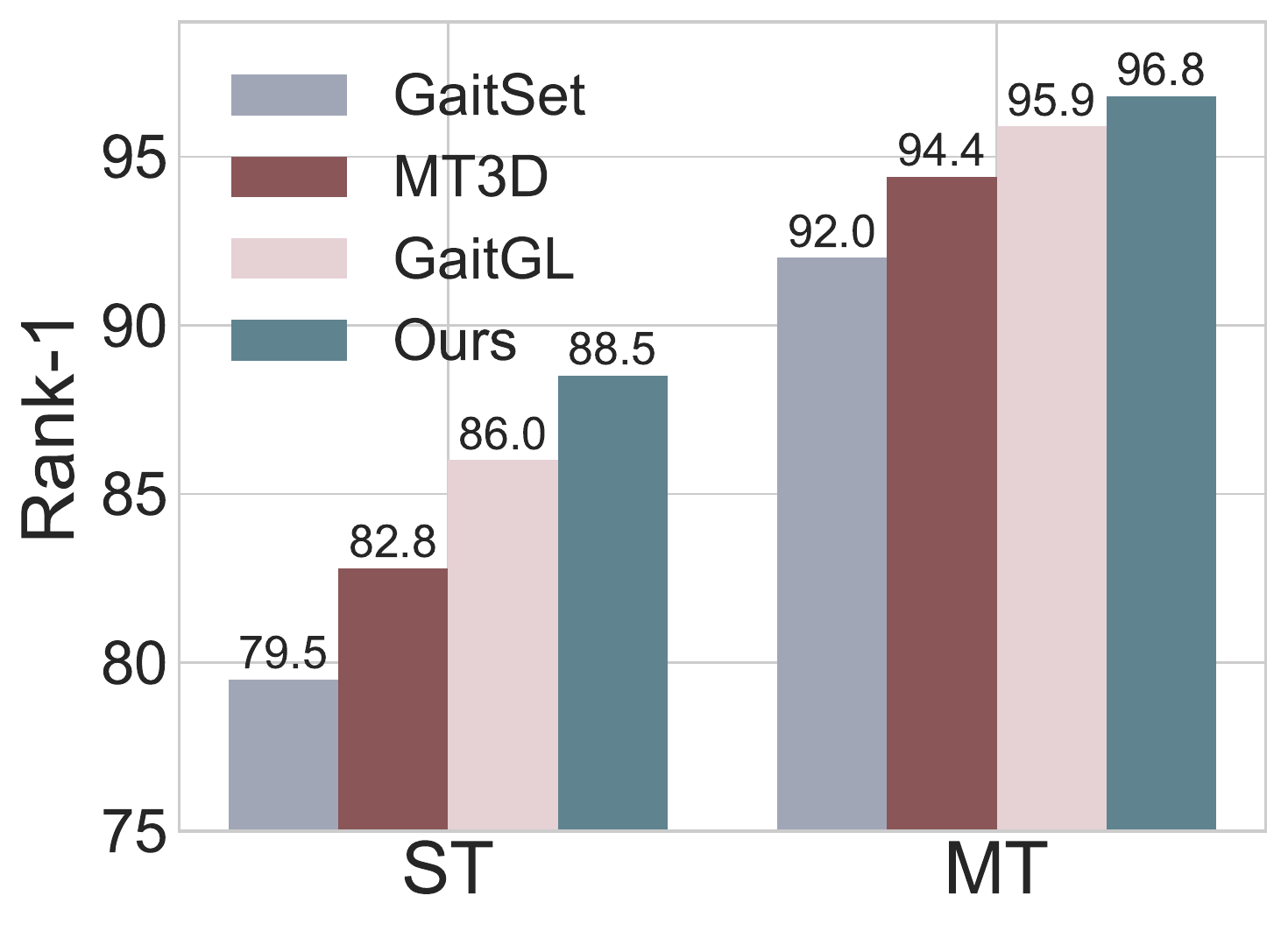}}
	\subcaptionbox{BG}{\includegraphics[width = 0.24\textwidth]{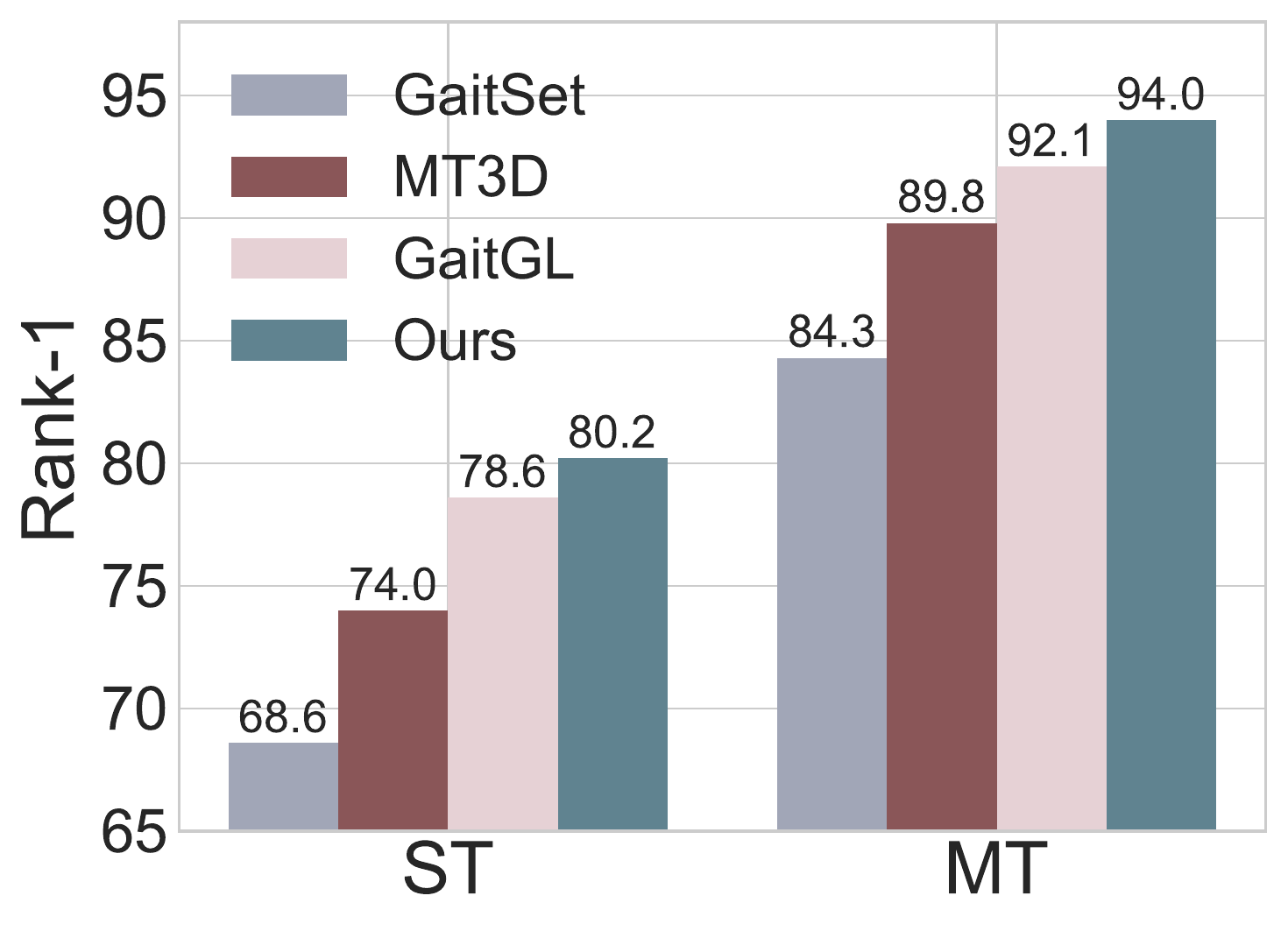}}
	\subcaptionbox{CL}{\includegraphics[width = 0.24\textwidth]{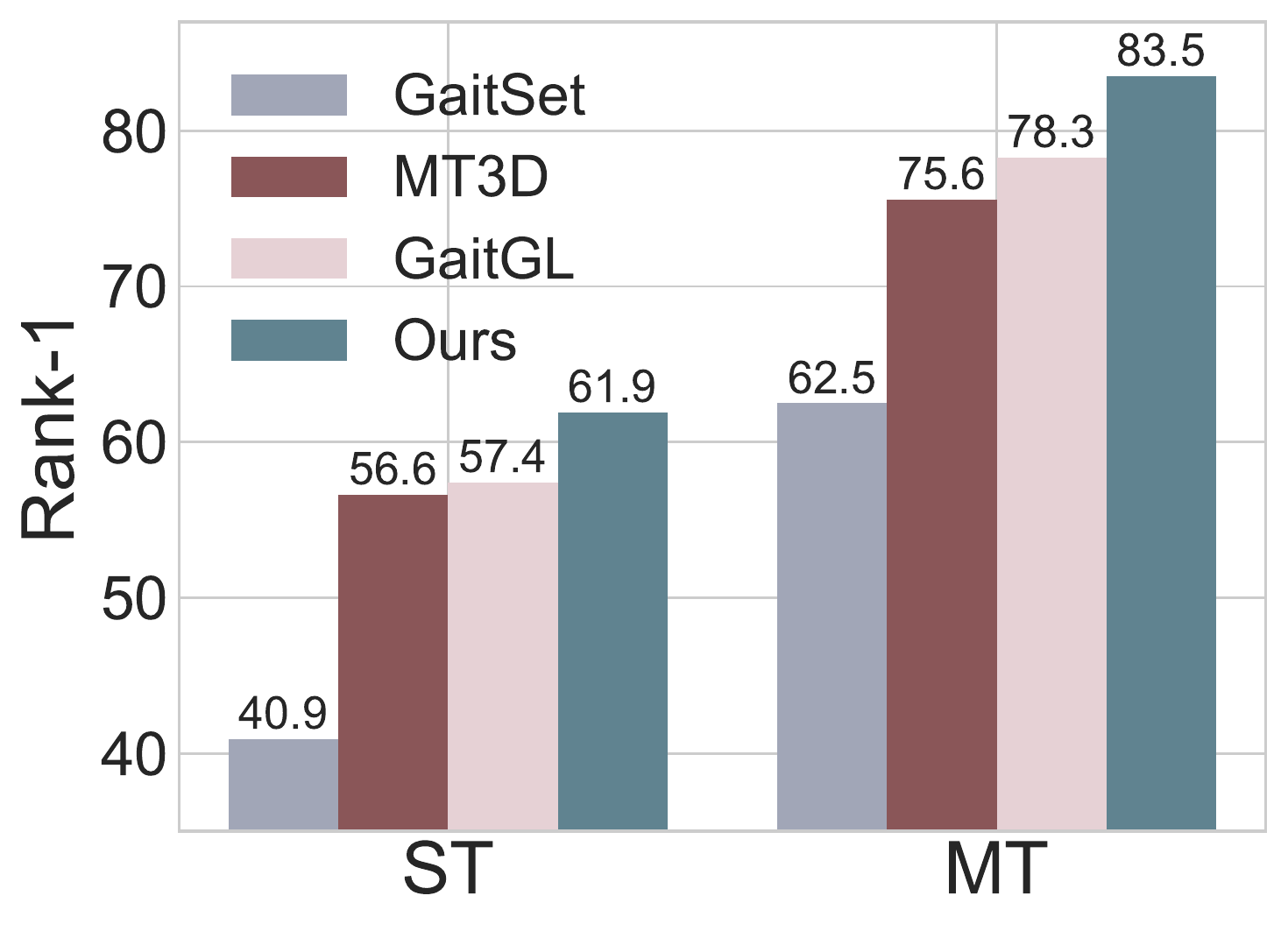}}
	\subcaptionbox{Mean}{\includegraphics[width = 0.24\textwidth]{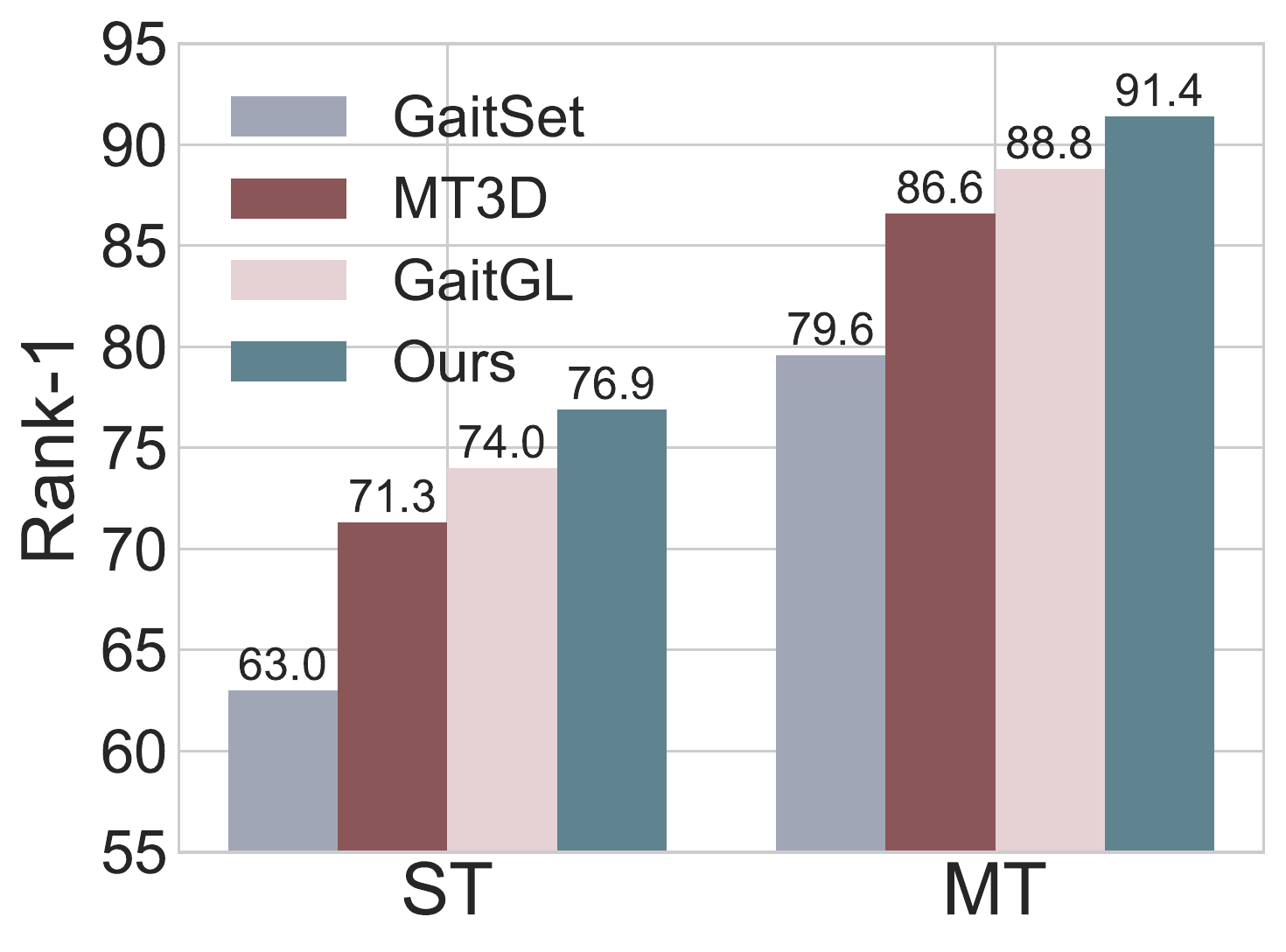}}	
\caption{Comparison with state-of-the-art methods under ST/MT setting.}
\label{fig:stmt}
\end{figure}

Then, we evaluate MetaGait under the data-limited scenarios following the protocol in~\cite{chao2019gaitset}.  As the experimental results are shown in~\cref{fig:stmt}, MetaGait outperforms state-of-the-art methods with a significant margin, which further shows the efficiency and robustness of MetaGait under small data scenarios.

\begin{table}[t]

    \centering
    \caption{Comparison with SOTA methods of rank-1 accuracy (\%) and mAP (\%).}
      \renewcommand{\arraystretch}{0.7}
    \setlength{\tabcolsep}{3mm}
    \begin{tabular}{c|c|cccc|c}
    \toprule
    \multirow{2}{*}{Method} & \multirow{2}{*}{Pub.} &  \multicolumn{4}{c|}{Rank-1} & \multirow{2}{*}{mAP} \\ \cmidrule{3-6} 
    & &  NM   & BG   & CL   & Mean &  \\
    \midrule
    GaitSet~\cite{chao2019gaitset}  & AAAI19 & 95.0 & 87.2 & 70.4 & 84.2&86.2 \\ 
    GaitPart~\cite{Fan_2020_CVPR} & CVPR20 & 96.2 & 91.5 & 78.7 & 88.8 & 88.7\\ 
    GLN~\cite{hou2020gait}      & ECCV20 & 96.9 & 94.0 & 77.5 & 89.5& 89.2 \\ 
    MT3D~\cite{Lin_2020_CVPR}     & ACM MM20   & 96.7 & 93.0 & 81.5 & 90.4&90.1 \\ 
    CSTL~\cite{Huang_2020_CVPR}     & ICCV21 & 97.8 & 93.6 & 84.2 & 91.9& - \\ 
    3DLocal~\cite{Huang_2021_ICCV2}  & ICCV21 & 97.5 & 94.3 & 83.7 & 91.8& -\\ 
    GaitGL~\cite{Lin_2021_ICCV}   & ICCV21 & 97.4 & 94.5 & 83.6 & 91.8& 91.5 \\ 
    \textbf{MetaGait}     &  -      &  \textbf{98.1}    &\textbf{95.2}      &  \textbf{86.9}   &  \textbf{93.4}   & \textbf{93.2}\\
    \bottomrule
    \end{tabular}
    \label{tab:avg}
    \end{table}

Further, to evaluate the comprehensive retrieval performance of MetaGait, we present the average rank-1/mAP performance in~\cref{tab:avg}, where mAP is computed by the reproduced methods. Specifically, MetaGait outperforms GaitSet by \textbf{9.2\%}/\textbf{7\%}, GaitPart by \textbf{4.6\%}/\textbf{4.5\%}, and GaitGL by \textbf{1.6\%}/\textbf{1.7\%}, which indicates the superior retrieval performance of MetaGait.

    \begin{table*}[t]
        \caption{Averaged rank-1 accuracy on OU-MVLP across different views excluding identical-view cases.}
        \centering
        \renewcommand{\arraystretch}{0.9}
    \setlength{\tabcolsep}{0.3mm}
        \scalebox{0.85}{
        \begin{tabular}{c|cccccccccccccc|c}
            \toprule
        \multicolumn{1}{c|}{\multirow{2}{*}{Method}} & \multicolumn{14}{c|}{Probe View}                                                                & \multirow{2}{*}{Mean} \\ \cmidrule{2-15}
        \multicolumn{1}{c|}{}                        & 0$^{\circ}$ & 15$^{\circ}$ & 30$^{\circ}$ & 45$^{\circ}$ & 60$^{\circ}$ & 75$^{\circ}$ & 90$^{\circ}$ & 180$^{\circ}$ & 195$^{\circ}$ & 210$^{\circ}$ & 225$^{\circ}$ & 240$^{\circ}$ & 255$^{\circ}$ & \multicolumn{1}{c|}{270$^{\circ}$} &                       \\ \hline
        GEINet                                       &   11.4&29.1&41.5&45.5&39.5&41.8&38.9&14.9&33.1&43.2&45.6&39.4&40.5&36.3&35.8                       \\
        GaitSet                                      &   79.5&87.9&89.9&90.2&88.1&88.7&87.8&81.7&86.7&89.0&89.3&87.2&87.8&86.2&87.1                     \\
        GaitPart                                     &   82.6&88.9&90.8&91.0&89.7&89.9&89.5&85.2&88.1&90.0&90.1&89.0&89.1&88.2&88.7                     \\
        GLN                                          &  83.8&90.0&91.0&91.2&90.3&90.0&89.4&85.3&89.1&90.5&90.6&89.6&89.3&88.5&89.2                       \\
        CSTL                                         &   87.1&91.0&91.5&91.8&90.6&90.8&90.6&89.4&90.2&90.5&90.7&89.8&90.0&89.4&90.2                      \\
        3DLocal                                      &   86.1&91.2&92.6&92.9&92.2&91.3&91.1&86.9&90.8&92.2&92.3&91.3&91.1&90.2&90.9                      \\
        GaitGL                                       &  84.9&90.2&91.1&91.5&91.1&90.8&90.3&88.5&88.6&90.3&90.4&89.6&89.5&88.8&89.7                      \\ \midrule
       MetaGait ($64\times44$)                                         &  88.2&92.3&93.0&93.5&93.1&92.7&92.6&89.3&91.2&92.0&92.6&92.3&91.9&91.1&91.9             \\ 
        \textbf{MetaGait ($128\times88)$}                                         &  \textbf{88.5}&\textbf{92.6}&\textbf{93.4}&\textbf{93.7}&\textbf{93.8}&\textbf{93.0}&\textbf{93.3}&\textbf{90.1}&\textbf{91.7}&\textbf{92.4}&\textbf{93.3}&\textbf{92.9}&\textbf{92.6}&\textbf{91.6}&\textbf{92.4}              \\
        \bottomrule    
        \end{tabular} }
        \label{tab:ou}
        \end{table*}

\noindent\textbf{Results on OU-MVLP.} To verify the effectiveness of MetaGait on the large dataset, we evaluate it on the largest public dataset OU-MVLP. As shown in~\cref{tab:ou}, it can be seen that MetaGait outperforms other SOTA methods by considerable margins, which proves the generalizability of MetaGait.

\subsection{Ablation Study}
This section presents ablation studies to validate the effectiveness of MTA and MTP, including the quantitative and qualitative analysis.

\noindent\textbf{Effectiveness of MTA and MTP.} The individual impacts of the MTA and MTP module are presented in~\cref{tab:abmain}. The baseline model refers to the feature extractor in~\cite{Lin_2021_ICCV} with traditional temporal aggregation (Max Pooling) and a separate FC layer. From the results, several conclusions are summarized as: 1) Using MTA or MTP individually can obtain \textbf{3.3\%} and \textbf{2.5\%} performance gain, respectively, which indicates the effectiveness of these modules. And MetaGait

\input{tab/main_local.tex}

\begin{table}[h]

\centering
\caption{Analysis of Meta Triple Attention, including the attention on three dimension, the calibration network, and the soft aggregation gate.}
\renewcommand{\arraystretch}{0.7}
    \setlength{\tabcolsep}{1mm}

\begin{tabular}{ccc|cc|c|cccc}
\toprule
\multicolumn{3}{c|}{Attention} &
  \multicolumn{2}{c|}{Calibration} &
  \multirow{2}{*}{Aggregation Gate} &
  \multirow{2}{*}{NM} &
  \multirow{2}{*}{BG} &
  \multirow{2}{*}{CL} &
  \multirow{2}{*}{Mean} \\ \cmidrule{1-5}
\multicolumn{1}{c|}{Spatial} &
  \multicolumn{1}{c|}{Channel} &
  Temporal &
  Static &
  Meta &
   &
   &
   &

   \\ \midrule
\multicolumn{1}{c|}{} & \multicolumn{1}{c|}{} &  &  & & & 96.8 & 93.8  & 84.0  & 91.5 \\
\multicolumn{1}{c|}{\checkmark} & \multicolumn{1}{c|}{} &  & \checkmark & & & 97.4 & 94.0  & 84.5  & 92.0  \\
\multicolumn{1}{c|}{} & \multicolumn{1}{c|}{\checkmark} &  & \checkmark &  && 97.0 & 94.1 & 84.1 & 91.7 \\
\multicolumn{1}{c|}{} & \multicolumn{1}{c|}{} & \checkmark & \checkmark &  && 96.8 & 94.0   &84.7  &  91.8 \\ \midrule
\multicolumn{1}{c|}{\checkmark} & \multicolumn{1}{c|}{\checkmark} &  &\checkmark  & &&  97.5 & 94.2  & 84.8  & 92.2 \\
\multicolumn{1}{c|}{\checkmark} & \multicolumn{1}{c|}{} &\checkmark  & \checkmark & && 97.6 & 94.3 & 85.0 & 92.3 \\
\multicolumn{1}{c|}{} & \multicolumn{1}{c|}{\checkmark} &\checkmark  & \checkmark &  && 97.1 & 94.1  & 84.9  & 92.0 \\ \midrule
\multicolumn{1}{c|}{\checkmark} & \multicolumn{1}{c|}{\checkmark} & \checkmark & \checkmark &  & &97.7  & 94.5 & 85.2  &  92.5  \\
\multicolumn{1}{c|}{\checkmark} & \multicolumn{1}{c|}{\checkmark} & \checkmark &\checkmark  &  &\checkmark&97.8  & 94.8 & 86.0  &  92.9  \\
\multicolumn{1}{c|}{\checkmark} & \multicolumn{1}{c|}{\checkmark} & \checkmark &  & \checkmark &&98.0  & 95.0 & 86.4 &  93.1  \\
\multicolumn{1}{c|}{\checkmark} & \multicolumn{1}{c|}{\checkmark} &\checkmark  &  & \checkmark &\checkmark&\textbf{98.1} & \textbf{95.2}  & \textbf{86.9} & \textbf{93.4}  \\ \bottomrule
\end{tabular}
\label{tab:attention}
\end{table}

\noindent  improves the performance by \textbf{4.4\%}; 2) Both MTA and MTP significantly improve the performance under the most challenging condition (\textit{i.e.,} CL) by \textbf{5.1\%} and \textbf{3.7\%}. 3) The performance gain with MTP is mainly reflected in the BG/CL condition, where temporal aggregation would be more crucial~\cite{Huang_2021_ICCV}.

\noindent\textbf{Receptive Field in Omni-scale Representation.} In MTA, we use the re-weighted combination of receptive fields in diverse scales to achieve omni-scale representation. To explore the effects of the different combinations, we use the convolutions with the kernel size of 1,3,5,7,9 in the local calibration network of MTA as shown in~\cref{tab:branch}. It can be seen that the performance is improved with the increase of the receptive field scale until the combination of \{1,3,5\}. In contrast, a larger receptive field decreases the performance, which may lie in that larger and more diverse receptive fields could improve the ability of feature representation, but the over-parameterized convolution is hard to optimize.

\noindent\textbf{Analysis of MTA.} To evaluate the effectiveness of MTA, we analyze it from three aspects, \textit{i.e.,} the attention design, the kind of the calibration network, and the soft aggregation gate. From the results shown in~\cref{tab:attention}, we could conclude: 1) MTA effectively improves the performance either using alone or in dimension combination;  2) The calibration network and the soft aggregation gate, which are parameterized by the meta-knowledge, clearly improve the rank-1 accuracy by \textbf{1.2\%}. The above experimental results indicate that our MHN can effectively improve the model's adaptiveness.

\begin{table}[t]
\centering
\caption{The ablation study on Meta Temporal Aggregation. }
\renewcommand{\arraystretch}{0.7}
    \setlength{\tabcolsep}{2mm}
\begin{tabular}{ccc|cc|cccc}
\toprule
\multicolumn{3}{c|}{Aggregation} & \multicolumn{2}{c|}{Weight Network} & \multirow{2}{*}{NM} & \multirow{2}{*}{BG} & \multirow{2}{*}{CL} & \multirow{2}{*}{Mean} \\ \cmidrule{1-5}
Max       & Mean      & GeM      & \multicolumn{1}{c}{Static}        & Meta                            &                     &                     &                       \\ \midrule
     \checkmark     &           &          &    --           &       --         &         97.5  & 94.2   &  85.4  & 92.3    \\
     
          &  \checkmark          &         &     --          &       --         &        96.3             &   93.4                  &   84.0                  &   91.2                    \\
          
          &           &   \checkmark       &      --        &   --             &          97.3           &         94.3             &     85.6                &      92.4                 \\ \midrule
          
    \checkmark        &      \checkmark       &          &    \checkmark           &        & 97.3              &      94.2               &        85.7          &            92.4            \\
    
          &    \checkmark         &   \checkmark         &   \checkmark            &                &      97.6              &      94.5              &        85.7          &            92.6           \\
          
     \checkmark       &           &     \checkmark       &    \checkmark       &     &    97.7            &            94.5         &       85.9              &       92.7                                   \\ \midrule
      \checkmark     &      \checkmark      & \checkmark          &  \checkmark              &                &   97.9                  &    94.7                 &    86.2                &         92.9              \\
      \checkmark    &      \checkmark      &   \checkmark        &               &     \checkmark            &       \textbf{98.1} & \textbf{95.2}  & \textbf{86.9} & \textbf{93.4}              \\ \bottomrule 
\end{tabular}
\label{tab:aggregation}
\end{table}

\noindent\textbf{Analysis of MTP.} The results in~\cref{tab:aggregation} shows the impacts of different temporal aggregation methods and weighting network. It can be seen that: 1) Different temporal aggregation methods used together provide performance gain by \textbf{0.6\%}. 2) Weighting network with meta-knowledge could effectively integrate the merits of three aggregation methods than the static one, which indicates that MTP could achieve more comprehensive and discriminative representation.

\begin{figure}[t]
\centering
	\subcaptionbox{Baseline}{\includegraphics[width = 0.15\textwidth]{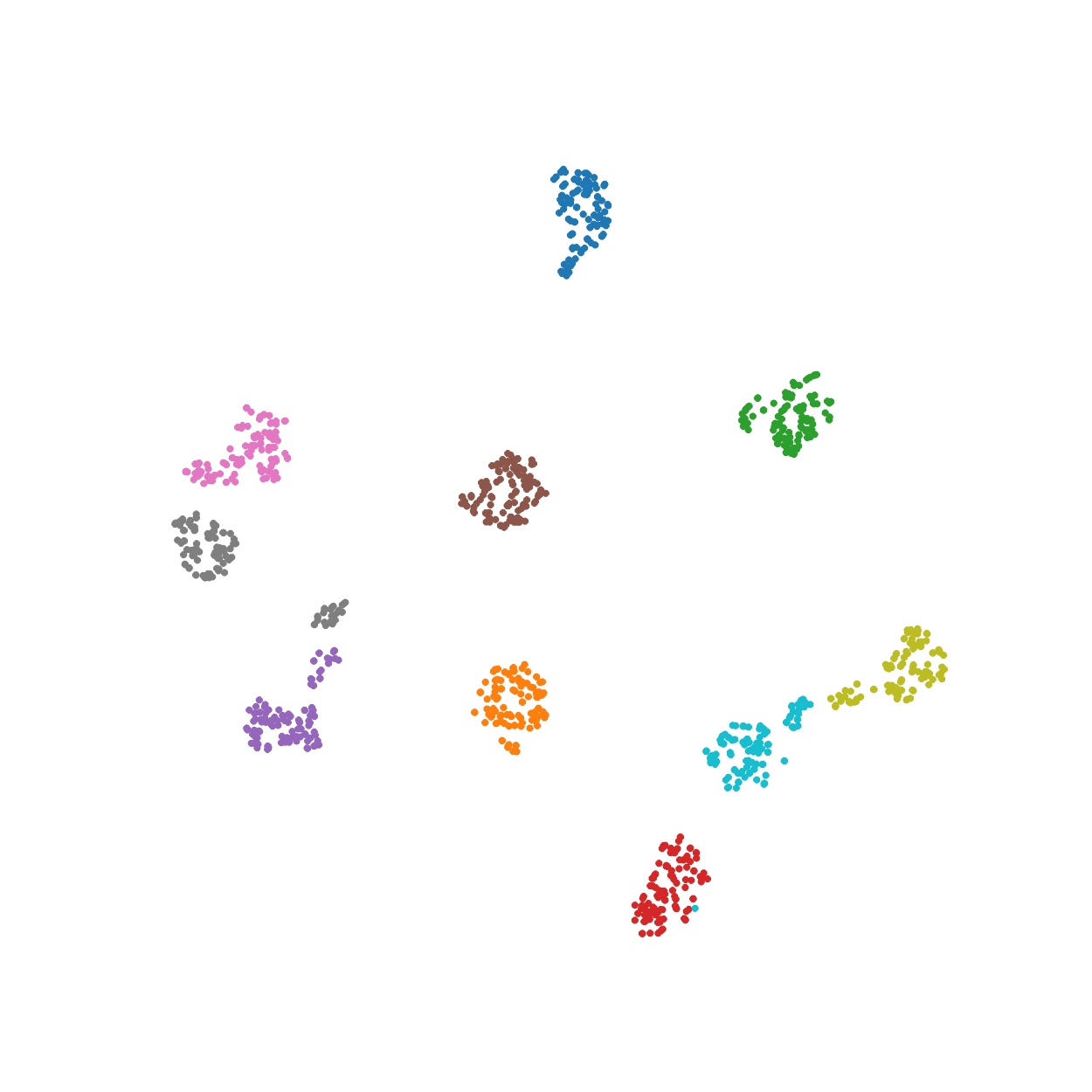}}
	\hspace{2.5cm}
	\subcaptionbox{Ours}{\includegraphics[width = 0.15\textwidth]{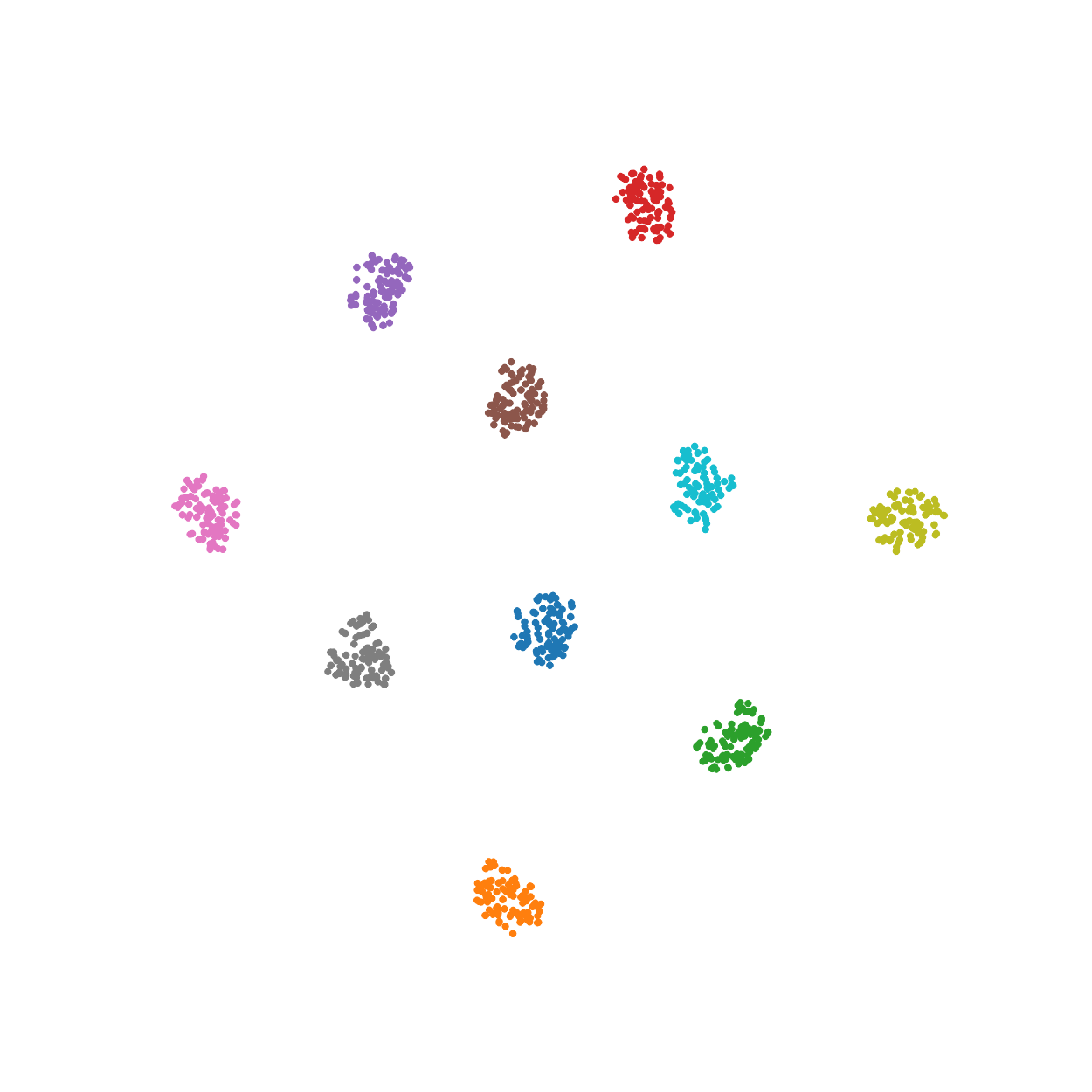}}
	\hfill
\caption{The visualizaton of feature space using t-SNE~\cite{van2008visualizing}.}
\label{fig:tsne}
\end{figure}

\noindent\textbf{Visualization of Feature Space.} To validate the effectiveness of MetaGait intuitively, we randomly choose 10 IDs from CASIA-B to visualize their feature distribution. As shown in~\cref{fig:tsne}, we find that MetaGait improves the intra-class compactness and inter-class separability than baseline.

\begin{figure}[t]
\centering
	\subcaptionbox{Sampled silhouette sequence from CASIA-B}{\includegraphics[width = 0.65\textwidth]{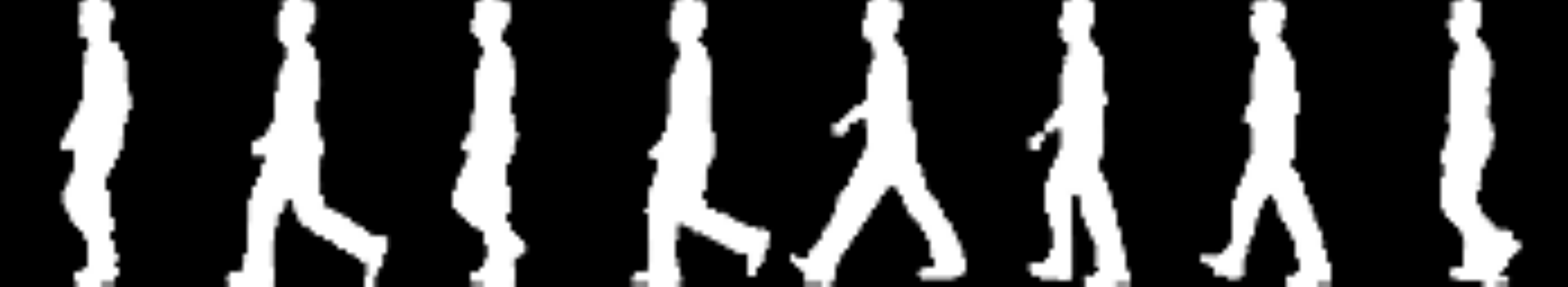}}
	\subcaptionbox{Meta Triple Attention on spatial dimension.}{\includegraphics[width = 0.65\textwidth]{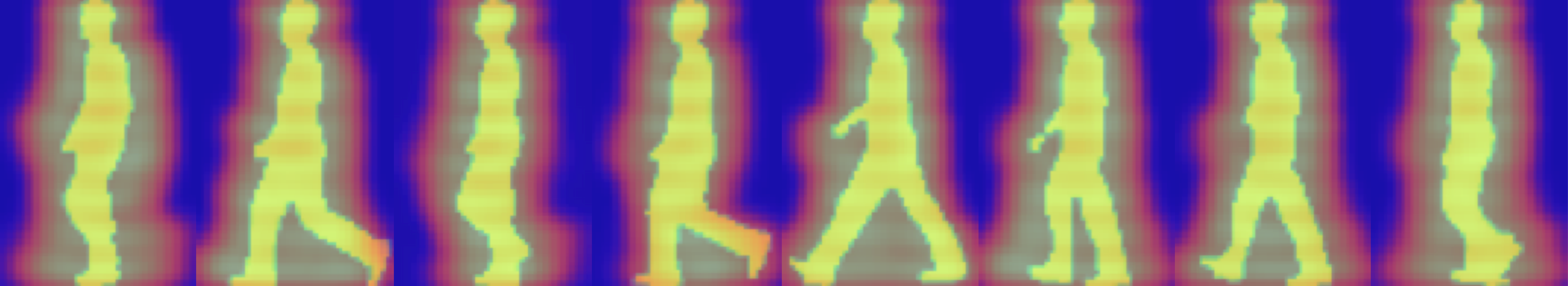}}
	\subcaptionbox{Meta Triple Attention on temporal dimension.}{\includegraphics[width = 0.65\textwidth]{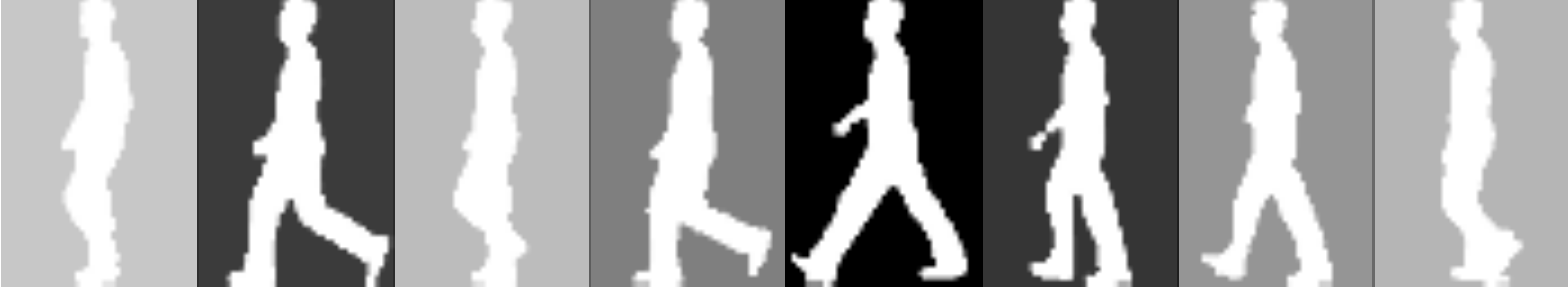}}
	\subcaptionbox{Meta Triple Attention on channel dimension.}{\includegraphics[width = 0.78\textwidth]{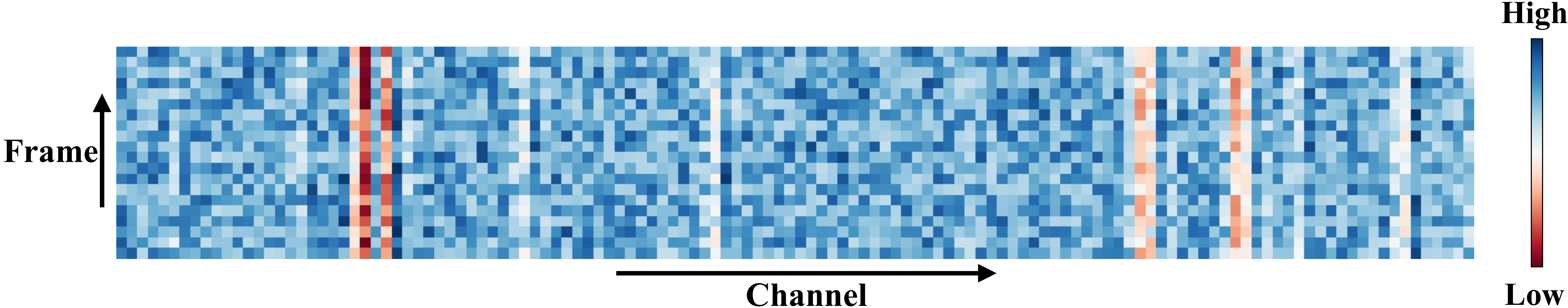}}
	\hfill
\caption{The visualization of the attention maps of Meta Triple Attention. The transparency of the silhouette in (c) represents its attention value.}
\label{fig:visual}
\end{figure}

\noindent\textbf{Visualization of Attention Maps.} To qualitatively analyze MTA, we visualize the attention map shown in~\cref{fig:visual}. For spatial dimension, MTA effectively learns the shape-aware attention map to guide the learning process adaptively. For temporal dimension, MTA can adaptively highlight important frames and suppress irrelevant frames to model the temporal representation. For channel dimension, it can be observed that MTA can learn a sample adaptive representation. Further, we can observe that different samples have low attention weights in certain channels, which may be caused by the channel redundancy in common.

\section{Conclusion}
We propose a novel MetaGait framework to alleviate the conflicts between limited visual clues and various covariates with diverse scales. The key idea is to leverage meta-knowledge learned from Meta Hyper Network to improve the adaptiveness of attention mechanism. Specifically, Meta Triple Attention utilizes meta-knowledge to parameterize the calibration network and simultaneously conduct omni-scale attention on spatial/channel/temporal dimensions. Further, Meta Temporal Pooling excavates the relation between three complementary temporal aggregation methods and aggregates them in a sample adaptive manner. Finally, extensive experiments validate the effectiveness of MetaGait.
\section*{Acknowledgements}
This work is supported in part by the National Natural Science Foundation of China under Grant U20A20222, National Key Research and Development Program of China under Grant 2020AAA0107400, Zhejiang Provincial Natural Science Foundation of China under Grant LR19F020004, NSFC (62002320, U19B2043) and the Key R\&D  Program of Zhejiang Province, China (2021C01119).

\clearpage
%
%
\bibliographystyle{splncs04}
\bibliography{main}
\end{document}